\documentclass[sigplan,screen,authorversion]{acmart}
\usepackage{preamble}
\usepackage{multirow}
\usepackage{enumitem}
\usepackage{microtype}
\usepackage[utf8]{inputenc}
\usepackage[T1]{fontenc}
\setitemize{noitemsep,topsep=0pt,parsep=0pt,partopsep=0pt}

\AtBeginDocument{%
  \providecommand\BibTeX{{%
    \normalfont B\kern-0.5em{\scshape i\kern-0.25em b}\kern-0.8em\TeX}}}

\setcopyright{rightsretained}
\acmPrice{}
\acmDOI{10.1145/3578360.3580273}
\acmYear{2023}
\copyrightyear{2023}
\acmSubmissionID{cc23main-p74-p}
\acmISBN{979-8-4007-0088-0/23/02}
\acmConference[CC '23]{Proceedings of the 32nd ACM SIGPLAN International Conference on Compiler Construction}{February 25--26, 2023}{Montréal, QC, Canada}
\acmBooktitle{Proceedings of the 32nd ACM SIGPLAN International Conference on Compiler Construction (CC '23), February 25--26, 2023, Montréal, QC, Canada}
\received{2022-11-10}
\received[accepted]{2022-12-19}

\definecolor{olivegreen}{RGB}{0,153,0}
\definecolor{awesome}{rgb}{1.0, 0.13, 0.32}
\mdfdefinestyle{mystyle}{
    backgroundcolor=white!20
}
\definecolor{ygreen}{HTML}{d6e39d}
\definecolor{sgreen}{HTML}{bee69c}
\definecolor{arylideyellow}{rgb}{0.91, 0.84, 0.42}
\definecolor{buff}{rgb}{0.94, 0.86, 0.51}
\definecolor{citrine}{rgb}{0.89, 0.82, 0.04}

\usepackage{letltxmacro}
\newif\ifcorrectingmode
\correctingmodetrue
\LetLtxMacro\origcite\cite

\renewcommand{\cite}[2][]{%
  \ifcorrectingmode
  \mbox{\origcite[#1]{#2}}%
  \else
  \origcite[#1]{#2}%
  \fi
}

\def \editMode {True}

\ifx\editMode\undefined
    \newcommand{\del}[1]{}
    \newcommand{\fixme}[1]{}
    \newcommand{\urkfixme}[1]{}
    \newcommand{\vk}[1]{}
    \newcommand{\ak}[1]{}
    \newcommand{\ra}[1]{}
    \newcommand{\sj}[1]{}
    \newcommand{\albert}[1]{}
    \newcommand{\fixmecleanuppl}[2][]{}
\else
    \newcommand{\del}[1]{}
    \newcommand{\fixme}[1]{\textcolor{red}{\textbf{URK:$\bigstar$}#1}}
    \newcommand{\urkfixme}[1]{\textcolor{red}{\textbf{URK FIXME}#1}}
    \newcommand{\vk}[1]{\textcolor{blue}{\textbf{VK:$\bigstar$}#1}}
    \newcommand{\ak}[1]{\textcolor{cyan}{\textbf{AK:$\bigstar$}#1}}
    \newcommand{\ra}[1]{\textcolor{olivegreen}{\textbf{RA:$\bigstar$}#1}}
    \newcommand{\sj}[1]{\textcolor{orange}{\textbf{SJ:$\bigstar$}#1}}
    \newcommand{\albert}[1]{\textcolor{brown}{\textbf{Albert:$\bigstar$}#1}}
    \newcommand{\fixmecleanuppl}[2][]{\fixme{CLEAN THIS UP PLEASE}}
\fi

\newcommand{\toolname}{{\sc RL4ReAl}\xspace}
\newcommand{\regalloc}{{regalloc}\xspace}
\newcommand{\rlforreal}[1][]{{\sc RL4ReAl}\xspace}
\newcommand{\rlforreall}[1][]{{\sc RL4ReAl-L}\xspace}
\newcommand{\rlforreallg}[1][]{{\sc RL4ReAl-G}\xspace}

\newcommand{\ragreedy}[1][]{{\sc Greedy}\xspace}
\newcommand{\rabasic}[1][]{{\sc Basic}\xspace}
\newcommand{\rapbqp}[1][]{{\sc PBQP}\xspace}
\newcommand{\rafast}[1][]{{\sc Fast}\xspace}

\newcommand{\mlregalloc}{{\texttt{MLRegAlloc}}\xspace}
\newcommand{\llvmgrpc}{{LLVM-g\sc{RPC}\xspace}}
\newcommand{\mirtovec}{{\sc MIR2Vec}\xspace}

\begin{document}

\title{\toolname{}: Reinforcement Learning for Register Allocation}

\author{S. VenkataKeerthy}
\orcid{1234-5678-9012}
\affiliation{
  \institution{IIT Hyderabad}
  \country{India}
}

\author{Siddharth Jain}
\orcid{0000-0003-3801-7759}
\affiliation{
  \institution{IIT Hyderabad}
  \country{India}
}

\author{Anilava Kundu}
\orcid{0000-0002-1275-8607}
\affiliation{
  \institution{IIT Hyderabad}
  \country{India}
}

\author{Rohit Aggarwal}
\orcid{0000-0001-8047-1184}
\affiliation{
  \institution{IIT Hyderabad}
  \country{India}
}

\author{Albert Cohen}
\orcid{0000-0002-8866-5343}
\affiliation{%
  \institution{Google}
  \country{France}
  }

\author{Ramakrishna Upadrasta}
\orcid{0000-0002-5290-3266}
\affiliation{%
  \institution{IIT Hyderabad}
  \country{India}
}

\renewcommand{\shortauthors}{S. VenkataKeerthy, S. Jain, A. Kundu, R. Aggarwal, A. Cohen and R. Upadrasta}

\begin{abstract}


We aim to automate decades of research and experience in register allocation, leveraging machine learning. We tackle this problem by embedding a multi-agent reinforcement learning algorithm within LLVM, training it with the state of the art techniques. We formalize the constraints that precisely define the problem for a given instruction-set architecture, while ensuring that the generated code preserves semantic correctness. We also develop a gRPC based framework providing a modular and efficient compiler interface for training and inference. Our approach is architecture independent: we show experimental results targeting Intel x86 and ARM AArch64. Our results match or out-perform the heavily tuned, production-grade register allocators of LLVM.
\end{abstract}

\ccsdesc[500]{Software and its engineering~Compilers}
\ccsdesc[500]{Computing methodologies~Multi-agent reinforcement learning}
\ccsdesc[300]{Mathematics of computing~Graph coloring}
\ccsdesc[500]{Computing methodologies~Multi-agent systems}

\keywords{Register Allocation, Reinforcement Learning}

\maketitle

\section{Introduction}

Register allocation is one of the well-studied and important compiler optimization problems. It involves assigning a finite set of registers to an unbounded set of variables. Its decision problem is reducible to graph coloring, which is one of the classical NP-Complete problems~\cite{Garey-Johnson-DBLP:books/fm/GareyJ79,bouchez2006register}.  Register allocation as an optimization involves additional sub-tasks, more than graph coloring itself~\cite{bouchez2006register}.
Several formulations have been proposed that return exact, or heuristic-based solutions.

Broadly, solutions are often formulated as constraint-based optimizations \cite{lozano2012constraint,kuchcinski2003constraints}, ILP~\cite{Appel2001,BarikGGPU06,chang1997,nagarakatte2007register}, PBQP~\cite{kim2020}, game-theoretic approaches~\cite{Pereira-PLDI-08-10.1145/1379022.1375609}, and are fed to a variety of solvers. 
In general, these approaches are known to have scalability issues.
On the other hand, heuristic-based approaches have been widely used owing to their scalability: reasonable solutions for practical benchmarks in near linear time. However, developing good heuristics is highly non-trivial and requires specialized domain expertise, on compiler construction as well as on hardware architecture. 
Various heuristics have been proposed over the past 40 years~\cite{chaitin1981,chow1990,briggs1994},  extending to recent times~\cite{chen2018}.
They are often fine-tuned for a particular architecture and yield non-optimal performance.

Recently, with the wide range of successes of Machine Learning (ML), ML-based approaches are being proposed to solve compiler optimization problems that have been known to be computationally expensive.
These include classical optimizations like phase ordering ~\cite{Fursin2011,MiCOMP, Jain-ISPASS-22-DBLP:conf/ispass/JainAVU22}, 
vectorization~\cite{hajali2019neurovectorizer}, 
function inlining~\cite{Simon-Inlining-10.1109/CGO.2013.6495004}, 
throughput prediction~\cite{IthemalMendis19a,Granite}. However, the applicability and effectiveness of ML methods to compiler optimizations under hard semantic constraints remains poorly understood. 
Focusing on register allocation (\regalloc{}), we identified some of the main reasons.

\begin{itemize}
    
    \item Regalloc is a complex problem, composed of multiple sub-tasks, including splitting, coalescing, spilling. These sub-tasks have to be  considered in addition to modeling hardware complexities.
    
    \item ML-based allocation schemes should ensure correctness: no two variables in the same live range be assigned to the same register, and the register types should be respected. Such semantic constraints should not suffer any approximation, unlike forgetful optimizations like function inlining.
  
    \item On a practical note, it is hard to integrate ML models and Reinforcement Learning (RL) algorithms in Python with compiler frameworks in C++ that are among the most complex pieces of software engineering.

\end{itemize}

Some initial attempts at addressing these challenges include
Das et al.~\cite{das2020} proposing a partially ML based solution,
Kim et al.~\cite{kim2022pbqp} leveraging RL to reduce the search space of PBQP, and the infrastructural
Compiler-Gym~\cite{cummins2021compilergym} approach to ease the RL-training process.

We propose a retargetable Reinforcement Learning ({\sc RL}) approach to the REgister ALlocation ({\sc REAL}) problem.
We formulate a \textit{multi-agent hierarchical reinforcement learning} optimization
considering program-specific information:
(1) to model the sub-tasks of register allocation like coloring, live range splitting and spilling, and 
(2) to encode the correctness constraints for preserving the semantics and hardware-compatible register assignments.
The legality of the register allocations and assignments is preserved by imposing constraints on the action space, or outcome of each agent.  As register allocation is a combinatorial problem, establishing the ground truth is hard, making RL a natural choice. It also facilitates the imposition of correctness constraints.

We leverage the LLVM infrastructure~\cite{Lattner:2004:llvm} to build the first end-to-end RL framework addressing the above-mentioned challenges. The interference graph of a function is extracted from the Machine Intermediate Representation (MIR) of LLVM. Instructions within each node are represented as vectors using representation learning methods. For this purpose, we propose MIR2Vec embeddings to represent MIR entities. These embeddings represent vertices of the interference graph that is traversed by RL agents. \mirtovec{} embeddings are application-independent and may be used for other backend applications in the future.
Finally, we propose \llvmgrpc{}, a generic framework to facilitate communication between the RL model and the compiler during training and deployment.
Our approach is portable: we show results on both Intel x86 and ARM AArch64. 

\paragraph{Contributions}
The following are our contributions:
\begin{itemize}
    \item The \emph{first end-to-end application} of RL for solving the register allocation problem.
    \item Formalizing the constraints to restrict the action space and preserve semantic correctness.    
    \item Proposal of \mirtovec{} to encode ML representations at Machine IR (MIR) level. 
    \item Design and implementation of \llvmgrpc{} for compiler integration with ML models.
    \item Experimental evaluation targeting x86-64 and AArch64 on SPEC CPU 06 and 17 benchmark suites.
\end{itemize}

\section{Background and Mathematical Model}
\label{sec:background}
We formulate the register allocation problem by defining  different constraints. We also give an overview of LLVM register allocators and multi-agent RL.

\begin{figure*}[tb]
    \begin{minipage}[b]{.495\textwidth}
    \begin{minipage}{.31\textwidth}
    \begin{lstlisting}[frame=single, language=c, label={Lst:src}, captionpos=b, escapechar=|]
    // Source
    i = 0
    x = 10
    y = 20 
    print x
    z = y / x 
    i++
    z = z + 10
    i++
    print y
    print z
    print i
    \end{lstlisting}
    \end{minipage}
    \hspace{0.2cm}
    \begin{minipage}{.55\textwidth}
    \begin{lstlisting}[frame=single, language=llvmMIR,label={Lst:MIR}, captionpos=b, escapechar=|, numbers=none]
MOV32ri 0, %i:gr32
MOV32ri 10, %x:gr32
MOV32ri 20, %y:gr32
<call print on %x>
$eax = COPY %y:gr32
<clear $edx>
IDIV32r %x:gr32, implicit-def $eax, implicit-def $edx
%z:gr32 = COPY $eax
%i:gr32 = ADD32ri %i:gr32, 1
...
<call print on %y, %z, %i>
    \end{lstlisting}
    \end{minipage}
    \caption{(a) Example source code and (b) its Machine IR 
    }
    \label{fig:code}
  \end{minipage}
  \begin{minipage}[b]{.5\textwidth}
    \hskip-17pt%
    \includegraphics[height=3.2cm]{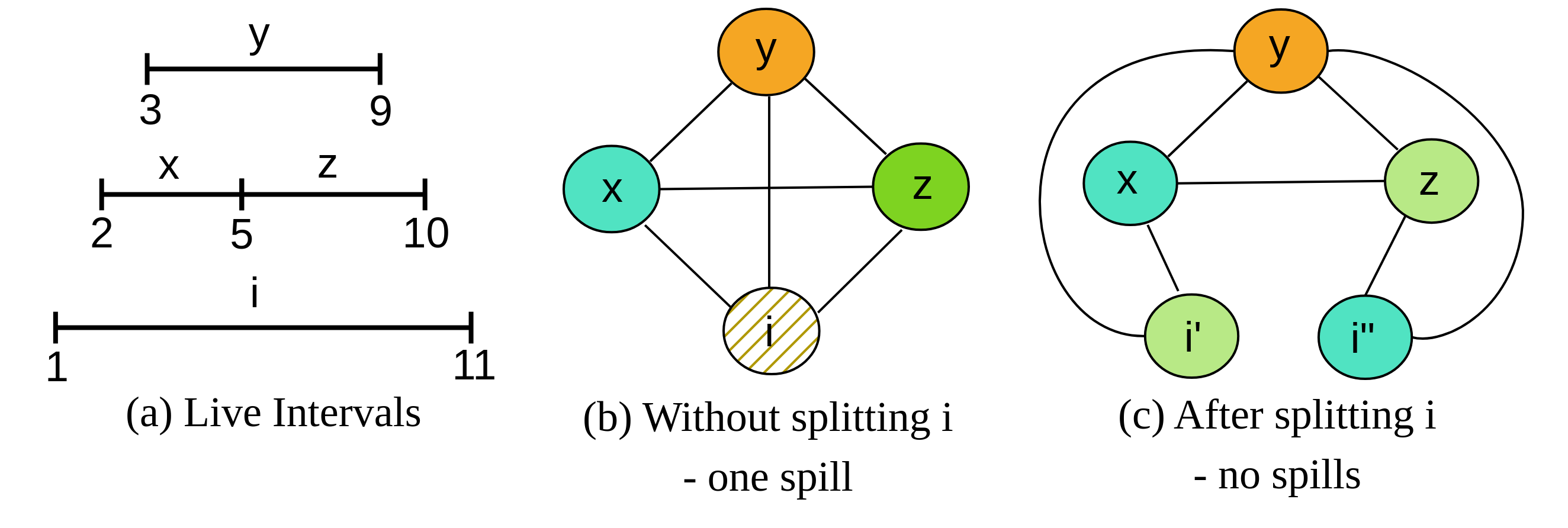}
    \caption{Register allocation with and without splitting}
    \label{fig:interference-graph}
  \end{minipage}

\end{figure*}

\subsection{Register constraints}
\label{subsec:constraints}
Optimizing compilers convert the source code into an Intermediate Representation (IR) where most target-independent optimizations take place. In the backend compiler, this IR is incrementally lowered to a machine-specific form.
This representation in LLVM is called as Machine IR (MIR). MIR at the stage of register allocation is very close to machine instructions, as instruction selection and other low-level optimizations have already been performed. 
After instruction selection, certain physical registers that are mandated by the architecture are immediately assigned. For instance, x86 processors mandate the output of 32-bit division to be stored only in \texttt{\$eax} and \texttt{\$edx} registers. 

As shown by the example in Fig.~\ref{fig:code}(a), \texttt{IDIV32} instruction divides the contents of \texttt{\$eax} and \texttt{\$edx} by \texttt{\%x} and stores the result in \texttt{\$eax}. Such mandatory assignments including calling conventions are made.
Regalloc can now be reduced to assigning physical registers to the other \textit{left-out} virtual registers ($\mathcal{V}$) while respecting the following constraints.

\paragraph{Type constraint}
The register file ($\textbf{R}$) of a machine consists of a collection of registers $R^t$ belonging to different \textit{types} ($t$): $\textbf{R} = \bigcup_tR^t$.
Assigning a physical register $r$ to a virtual register $v$ of type $t$, $v^t \blacktriangleright r$, should satisfy the register \textit{type} constraint $\chi^\mathcal{T}(v^t) = \{v^t \blacktriangleright r: r \in R^t\}$.
In Fig.~\ref{fig:code}(b), each virtual register is associated with a particular register type.
For instance, \texttt{\%x} is of \texttt{gr32} type, which means that it belongs to a 32-bit wide general-purpose register type. Meaning, only registers belonging to that type (like \texttt{\$eax}) can be assigned.

\paragraph{Congruence constraint}
Real-world instruction set architectures like x86 and AArch64 have a hierarchy of register \textit{classes}. 
For instance, 32 bit type of registers (like \texttt{\$eax, \$ebx}) are physically \textit{part of} the 64 bit ones (like \texttt{\$rax, \$rbx}).
We consider the registers that adhere to this \textit{part of} relation as a congruence class $\mathcal{C}(r)$.
For example, registers \texttt{\$al, \$ah, \$ax, \$eax, \$rax} of x86, which are ``chunks'' of the \textit{same physical register} belong to the same congruence class, satisfying $\mathtt{\$al}, \mathtt{\$ah} \sqsubseteq \mathtt{\$ax} \sqsubseteq \mathtt{\$eax} \sqsubseteq \mathtt{\$rax}$.
So, the assignments for virtual register $v$ should be among the set of registers that satisfy the following \textit{congruence} constraint $\forall r' \in \mathcal{C}(r) $:
\[
\chi^\mathcal{C}(v^t) = \{v_i^t\!\blacktriangleright\! r : \forall v_i, v_j \in \mathcal{V}, v_i \neq v_j, \nexists v_j \blacktriangleright r' \in L(v_i)   \}
\]

Here $L(v)$ corresponds to the live range of variable $v$, and is computed as $ L(v) = [P_v^{\textit{def}}, P_v^{\textit{end}}]$; the definition of $v$ occurs at program point $P^{\textit{def}}$, and its last use is in $P^{end}$.
Fig.~\ref{fig:code}(a) gives an example. The live ranges of the corresponding variables are shown in Fig.~\ref{fig:interference-graph}(a).

\paragraph{Interference constraint}
\label{subsec:regalloc}
Register allocation has been modeled as a graph coloring problem~\citep{chaitin1981}. 
For each function in the program, it involves creating an \textit{Interference graph} $G(V, E)$ defined as follows: the vertices of the graph are mapped to virtual registers ($v$) or physical registers ($R_a$), meaning $V \in (\mathcal{V} \cup R_a)$; the edges $E$ are computed as $\{(v_i, v_j): v_i, v_j \in V \wedge L(v_i) \cap L(v_j)\}$.
The interference graph corresponding to the example in Fig.~\ref{fig:code} is shown in Fig.~\ref{fig:interference-graph}(b). 
The \textit{interference} constraint says that no two adjacent nodes in $G$ should be allocated the same color.
The set of registers satisfying this constraint is given by:

\[
\chi^\mathcal{I} (v^t) = R^t \setminus \{r : \forall u((u, v)\in E \wedge \left ( u\blacktriangleright r) \right ) \}
\]

In summary, for a given virtual register $v$ of type $t$, the set of available registers for allocation $\chi(v^t)$ is defined as the set of registers that satisfy the (i) type, (ii) congruence, and (iii) interference constraints:

\smallskip
\centerline{%
$
\chi(v^t) = \chi^\mathcal{T}(v^t) \cap \chi^\mathcal{C}(v^t) \cap \chi^\mathcal{I} (v^t)
$
}

\subsection{Live range splitting and spilling}
\label{sec:splitting-spilling}

The above formulation of register allocation in terms of graph coloring is well known and natural, as a \emph{decision problem}. Yet register allocation as an \emph{optimization problem} is actually \textit{much more} than graph coloring.
For instance, when there are not enough physical registers available, deciding which variable (virtual register) has to be spilled to memory is important, as memory accesses take far more time than register accesses. 
Spilling a variable, $\mu(v)$ involves writing/reading it to/from a memory location on access.
A trivial example is the loop induction variable: it would incur high cost to read/write from/to memory, if a decision is made to spill it. Hence, register allocators try to reduce the spill cost $M(v)$, in addition to minimizing the number of spills.
For a  machine with 3 registers, the example code shown in Fig.~\ref{fig:code} is not \textit{3-colorable}, and results in spilling a variable. 

A live range of a virtual register can be split.
Let $k \in K$ denote a program point among the uses of $v$. 
Splitting live range of $v$ at $k$ is defined as $\varphi(v, k): L(v) \leadsto (L(v'), L(v'')); L(v') = [P_v^{def}, P_v^k], L(v'') = [P_v^{k+1}, P_v^{end}]$.
In Fig.~\ref{fig:interference-graph}(b), splitting $i$ at $5$ into ($i'$, $i''$) makes the graph \textit{3-colorable}.
Determining which variable to split, and at which point is non-trivial.

\subsection{Register allocators in LLVM}
\label{subsec:llvm-regalloc}

The LLVM compiler currently has 
four register allocators: \rafast{}, \rabasic{}, \ragreedy{}, and \rapbqp{}, ranked according to implementation complexity. They are implemented as passes and operate on one function at a time.

\rafast{} is an improved version of the linear scan algorithm 
\cite{poletto1999} and operates at the basic block level.  
\rabasic{} is an improved variation of \rafast, and operates at the function level \cite{xavier2012}. 
\ragreedy{} was developed~\cite{greedy-alloc-llvm} to address the shortcomings of \rabasic; it iteratively combines four strategies: splitting, spilling, coalescing (merging of live ranges), eviction (de-allocating the already-allocated physical register). 
Each of these strategies is driven by greedy heuristics. 
\ragreedy{} is a complex, highly tuned, default regalloc at \texttt{-O3} optimization level. It iterates over the virtual registers in multiple rounds and obtains a legal physical allocation if possible. 
It includes \textit{highly tuned heuristics} (i) for placing of proper spill code, (ii) to minimize the load-store 
instructions while favoring register moves by applying strategies like node-splitting/eviction/last-chance-recoloring, etc.
\rapbqp{} is the only solver based mechanism in LLVM, and models it as a Partitioned Boolean Quadratic Problem to obtain allocations~\cite{hames2006nearly}.
These allocators are a result of significant (\textit{man-decades}) amount of engineering by expert compiler writers; and are continually improved to address the regressions on a case-by-case basis.
Not all allocators implement all strategies; live range splitting only takes place in \ragreedy{}, whereas coalescing is present in \ragreedy and \rapbqp{}, and eviction is in iterative allocators like \ragreedy and not \rapbqp{}.

\subsection{Multi-Agent Hierarchical RL}
\label{subsec:marl}
Reinforcement learning (RL) is a branch of machine learning that often tries to solve the problems where enumerating ground truth is either hard or infeasible. The learning happens with \textit{experience} where the agent learns a policy to determine the best possible action based on its \textit{observation} from the \textit{environment}. Depending on the \textit{goodness} of action, the environment gives positive or negative rewards so as to course-correct the learning. With the evolution of DL, methods like PPO~\cite{schulman2017ppo} that use gradient based approaches to learn better policy have become prevalent.

Depending on the problem formulation, there can be single or multiple agents to solve the problem. When the problem is modeled with multiple agents, it is called \textit{Multi-Agent Reinforcement Learning} (MARL). If all the agents work together towards a common goal, learning is said to be cooperative. If they compete against each other to achieve a goal, the learning is said to be competitive. In certain cases, there can be a mix of both of these. MARL is an active field of research that has accomplished huge success in gaming~\cite{silver2017mastering}, robotics~\cite{kalashnikov2018qt}, navigation~\cite{almasan2020deep}, and autonomous driving problems~\cite{KiranRLSurvey2021}.

We formulate \regalloc{} as a number of smaller subproblems using multiple agents to  model node selection, task selection (among splitting, spilling, or coloring), splitting, and coloring. These agents work cooperatively to achieve a beneficial register allocation. 
Another categorization in the case of MARL problems is based on the schedule of the agents. If the agents act on the environment sequentially, it is the sequential variant of MARL. If the agents form a hierarchy, where the top-level agent determines how the agents at the lower level should act, the learning is said to be hierarchical. 
In \toolname{}, task selection, splitting, and coloring are modeled in a hierarchical fashion.

\section{Modeling \toolname{}}
\label{sec:model}
\begin{figure*}
    \centering
    \includegraphics[width=\textwidth]{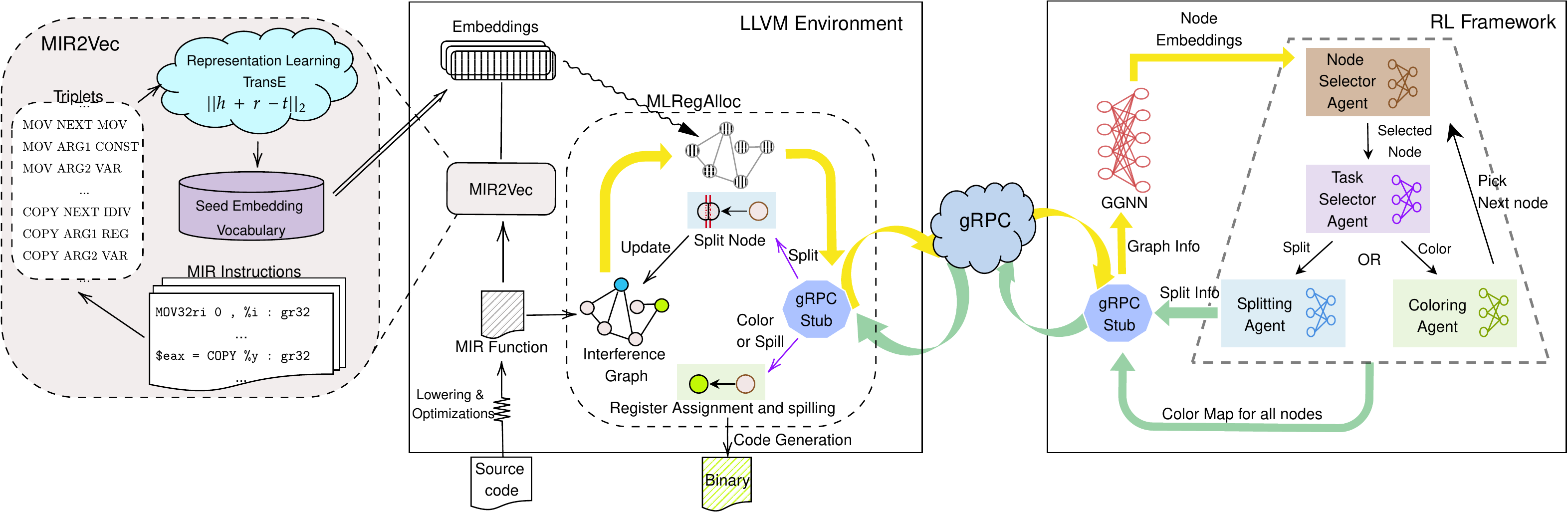}
    \caption{Overview of \toolname{}: The interference graph ($G$) generated from MIR is represented as \mirtovec{} vectors and is passed as input to the RL Framework via \llvmgrpc{} 
    The agents in the RL framework perform splitting/coloring on $G$, and the register assignments (colors) are communicated back to LLVM via \llvmgrpc{}. 
    In case of splitting a vertex in $G$, the split points predicted by the agent are passed to the LLVM environment to perform splitting, and the updated graph is sent back to the model as response.}
    \label{fig:overview}
\end{figure*}

We formulate register allocation as a Markov Decision Process (MDP) using hierarchical Reinforcement Learning (RL), modeling the sub-tasks of register allocation as lower level tasks controlled by multiple agents.
Fig.~\ref{fig:overview} sketches the overall approach. It involves interactions between the LLVM compiler and the RL model for both training and inference.

\subsection{Environment}
We implemented a new \mlregalloc{} pass in LLVM, to generate an interference graph ($G$), allocate, split and spill registers as predicted by the agents. This pass also generates a representation of $G$ using \mirtovec{}.

\subsection{Agents}
The task of allocating registers is split into multiple sub-tasks. Each of these tasks are modeled as agents $\{\omega_\upsilon, \omega_\tau, \omega_\varphi, \omega_\xi\}\in \Omega$, that learn their respective policies $\pi_\omega$ to optimally solve the low level tasks. We formulate hierarchical agents for four sub-tasks as shown in Fig.~\ref{fig:overview}: 
\begin{itemize}
    \item Node selector ($\omega_{\upsilon}$): Top level agent that learns to pick a node $v \in G$.
    \item Task selector ($\omega_{\tau}$): Mid level agent that learns to select a task among $\{\chi, \varphi\}$ on $v$ picked by $\omega_{\upsilon}$.
    \item Splitter ($\omega_{\varphi}$): Low level agent that learns to identify a split point $k$ for $v$.
    \item Coloring Agent ($\omega_{\xi}$): Low level agent that learns to pick a valid color $\chi_i \in \chi$ or spill $\mu$.
\end{itemize}

As it can be seen, each high level agent invokes a low level agent while following the timeline:  $\omega_\upsilon \prec \omega_\tau \prec \{\omega_\varphi, \omega_\xi\}$.
Each agent $\omega$ has its own state space $S_\omega$, action space $A_\omega$, and reward $R_\omega$ to learn a policy $\pi_\omega$.

\paragraph{Coloring Agent ($\omega_\xi$)}
In case of regular architectures like x86 and AArch64, all the registers of the same type/class have equal effect on the performance. 
However, the standard register allocators prefer some registers over others within the same class while considering several heuristics.
One of the predominantly used heuristic is to give preference to the physical registers that are not (aliases of) callee saved registers. 
This is achieved by deriving an ordering of physical registers that can be allocated to a virtual register while satisfying the constraints described in Sec.~\ref{subsec:constraints}.  The goal of this ordering is to reduce the number of register moves. 
Hence, we design a simple model that can act as a coloring agent to learn the beneficial color assignments.

For a graph of $V$ nodes and a set of registers 
$\chi(v)$
available at the instant, the state space of $\omega_\xi$ is given as a tuple $\langle \llbracket v \rrbracket, |\chi(v)|, |V_{\textrm{nclr}}| \rangle$, where $V_{\textrm{nclr}} = V \setminus V_{\textrm{clrd}}$ are the nodes to be colored, $v$ is the node that is picked by $\omega_\upsilon$, and $\llbracket.\rrbracket$ denotes its embedding. Meaning, the coloring agent uses the following information to decide the register to be assigned: the embedding of the vertex $v$, the number of registers satisfying the constraints (see Sec.~\ref{subsec:constraints}), and the number of uncolored nodes in $G$.  If no registers are available, the coloring agent marks $v$ for spilling. Hence the legal action space of $\omega_\xi$ is:
\[
A(\omega_\xi) =
\begin{cases}
        \chi(v),& |\chi(v)| > 0\\
        \mu(v),& \mathrm{otherwise}
\end{cases}
\]
$\chi(v)$ gives the set of legal registers for $v$ (Sec.~\ref{subsec:constraints}). To improve performance, the agent should maximize the use of registers for vertices with higher spill weight. And, spill weight roughly corresponds to the importance of the node $v$.  Hence the reward for the coloring agent is given as:
\[
    R(\omega_\xi) = 
\begin{cases}
    + M(v),& \text{if}\ \chi(v) \\
    - M(v),& \text{if}\ \mu(v)\\
\end{cases}
\]

In our experiments, spill weight $M$ is estimated using the \textit{spill costs} computed by LLVM.

\paragraph{Splitter ($\omega_\varphi$)}
Live range splitting $\varphi(v, k)$ corresponding to a variable $v$ involves inserting a \texttt{move} instruction at the split point $k$ and creating two new live ranges $L(v')$ and $L(v'')$  in place of a single original live range $L(v)$ as explained in Sec.~\ref{sec:splitting-spilling}. 
Selecting an appropriate split point plays an important role in making effective spilling and coloring decisions. For instance, the \ragreedy{} in LLVM \textit{greedily} selects the split points so as to carve out a region that can be allocated a register, while the other parts are spilled. In our model, the splitting agent predicts the split point in the live range among all the points of use.

Inserting the \texttt{move} instructions can be seen as a dataflow problem, 
that is analogous to the \texttt{phi} (\texttt{copy}) placement while creating (going out-of-) SSA form. 
We use dominance frontiers to place the \texttt{move} instructions appropriately to preserve the correctness.
In Algorithm~\ref{algorithm:move-placement}, we
show how the move instructions are inserted. 
This algorithm directly builds from earlier works~\cite{bouchez2006register, brisk2005polynomial, hack2006register} and dominance property~\cite{ssa:cytron1991efficiently}, so its soundness can easily be proved. 

\begin{algorithm}
\DontPrintSemicolon
\SetAlgoNoLine
\textbf{Parameter}: Virtual register $v$, Split point $k$\\
Rename  $v \rightarrow v'$\\
At use point $k$ do: $v'' \leftarrow$ \texttt{move}($v'$)\\
Basic block $B \leftarrow block(v_k)$\\
\For{$i \in DominanceFrontier(B)$}{
     $v' \leftarrow$ \texttt{move}($v''$), after last use($v'$) in $i$\\
     Rename $v' \rightarrow v'',\ \forall use(v')$ between $B$ and $i$
}

\caption{move-placement in live range splitting}
\label{algorithm:move-placement}
\end{algorithm}

For predicting where to split the live range of a variable $v$, the node splitter considers the spill weights at each use of the variable $\mathcal{M}(v) = \{M(v_k): \forall k \in K\}$, the distances between each successive use $D_v = \{D(v_i, v_{i+1}), \forall i, i+1 \in K\}$, and the embedding $\llbracket v \rrbracket$ of $v$. 
The use distance is the number of program points between two uses of $v$. 
Hence, the state space is given as a tuple $\langle \llbracket v \rrbracket, \mathcal{M}(v), D_v\rangle$.
For a given state, the agent learns an optimal program point $p \in K$ where $v$ can be split.
Hence the action space $A(\omega_\varphi) = K$.

The reward for the agent on splitting $v$ into $(v', v'')$ is based on the difference in spill weights of the variable before and after splitting. The agent gets a positive reward on reducing the sum of the spill weights, which indicate a reduction in complexity of coloring. 
\[
    R(\omega_\varphi) = M(v) - \sum_{i \in \{v', v''\}} M(i)
\]

\paragraph{Task Selector ($\omega_\tau$)}
For selecting a task ($\tau$) among coloring and splitting, the agent $\omega_\tau$ considers the parameters specific to each of the tasks: the representation of $v$, the number of available registers, the number of interferences, its life-time, spill weight. Hence the state space is formulated as the tuple:
$\langle \llbracket v \rrbracket, |\chi(v)|, \delta(v), |K(v)|, M(v) \rangle$.
The action space of $\omega_\tau$ is defined as:
\[
A(\omega_\tau) =
\begin{cases}
        \{\varphi, \chi\},& |K(v)| \geq k\\
        \chi,& otherwise
\end{cases}
\]
Here $|K(v)| \geq k$ indicates that $v$ should have at least $k$ uses to be considered for splitting. We define $k$ as a hyper-parameter. We set $k=2$ (1 definition and 1 use) in our experiments. We model the reward for this agent based on the outcome of the low level tasks. 

\[
R(\omega_\tau) = 
\begin{cases}
    R(\omega_\xi),& \tau = \chi \\
    0, &  \tau = \varphi
\end{cases}
\]
On choosing to color, the agent gets a reward based on coloring decision of $\omega_\xi$. However, if the agent chooses to split, we defer from giving a reward as the goodness of the outcome is not known till a coloring decision is made.

\paragraph{Node Selector ($\omega_\upsilon$)}
It is well known that the order of picking a variable for allocation would highly affect the final outcome of register allocation.
Usually, the iterative allocators use a priority queue to determine the order of allocation. The priority is derived from different heuristics including the spill cost of the variables, size of the live ranges, among others. 
We use a model ($\omega_\upsilon$) to figure out this order of allocation by predicting the variable/vertex to process (split/color) at every step of allocation after processing the previous vertex.

The state space of $\omega_\upsilon$ comprises the embedding of each vertex in $G$ obtained from a Gated Graph Neural Network (GGNN) as explained in Sec.~\ref{sec:representing-interference-graphs}.
Along with these embeddings, the agent uses the spill weights of the nodes $M$ to characterize the state.
Hence, the state space is given as a tuple $\langle \llbracket G \rrbracket, M(V) \rangle$
Its legal action space is $A(\omega_\upsilon) = V_{\textrm{nclr}}$.
The learned policy is deemed \textit{good} based on the final coloring decision of the node. 
Hence, the reward for this agent is also modeled based on the rewards of the coloring agent ($\omega_\xi$):
\[
R(\omega_\upsilon) = 
\begin{cases}
    R(\omega_\xi),& \tau = \chi \\
    0, & \tau = \varphi \\
\end{cases}
\]

\paragraph{Global rewards}
In addition to providing rewards to the agents at each step, we derive a global reward based on the throughput of the generated function estimated by LLVM-MCA~\cite{llvm-mca}.
The global reward is computed based on the difference between the throughputs of the code generated by \toolname{} ($Th_\toolname{}$) and \ragreedy{} ($Th_{Greedy}$) as:

\[
R_G = 
\begin{cases}
    +10,& Th_\toolname{} \ge Th_{Greedy}  \\
    -10, & Otherwise\\
\end{cases}
\]

This way of using throughput helps capturing the overall impact of the allocation scheme in comparison with \ragreedy{}.
\section{Representing Interference Graphs}
\label{sec:representing-interference-graphs}
We represent nodes of the interference graph as embeddings obtained from LLVM's MIR instructions.
Such embeddings form the input to a Gated Graph Neural Network (GGNN) that learns to generate the representation of the state space.

Any deep learning model can accept only a numerical representation as input. When it comes to applying ML techniques on programs, there are two possible ways: (i) feature based representations~\cite{Fursin2011}, or (ii) distributed representations/embeddings~\cite{ncc, VenkataKeerthy-2020-IR2Vec, flow2vec2020}. 

It is widely understood that program embedding techniques at IR-level automatically capture the semantic information that may be difficult to recover with only syntactic embeddings \cite{VenkataKeerthy-2020-IR2Vec}. Further, the IR-based program embeddings generalize better across applications, while effectively requiring much less data to train.

We propose \mirtovec{}, a learned embedding model for representing the MIR form of the program. The learned \mirtovec{} representations are in the form of $n$-dimensional real-valued vectors, which can be passed to the model for learning a downstream optimization task like register allocation. The embeddings can be seen as the key means for facilitating the current optimization problem (viz.\ regalloc), but, also a necessary means in which other backend problems (viz.\ instruction scheduling) can be easily modeled to obtain a \textit{(representation) learning based infrastructure} for backend optimizations. We generate \mirtovec{} representations by: creating triplets by forming relations between entities, training TransE~\cite{transe-Bordes:2013:TEM:2999792.2999923} to obtain the seed embedding vocabulary, and using it to create instruction-level representations as shown in Fig.~\ref{fig:overview}. As MIR is target-specific, the embeddings are also specific to the architecture.

\paragraph{MIR Entities}
Opcodes and MIR instruction arguments form the entities. Arguments primarily include physical and virtual registers, and immediate values. We abstract these arguments with generic identifiers as a preprocessing step.

We create two different relations.  (i) $NextInst$: Captures the relation between the current opcode and the next instruction opcode, (ii) $Arg_i$: Captures the relation between the opcode and the arguments of the instruction. Once the triplets are generated, we train the TransE model to obtain the embeddings for each of the entities. 

\paragraph{Grouping of opcodes}
MIR contains specialized opcodes, in terms of the operating width among other factors. MIR contains about 15.3K different possible opcodes in x86 and about 5.4K in AArch64. Obtaining a dataset to cover all such specialized operands would be highly infeasible, and in turn, would not generate good representations. 
Hence we mask out the opcodes based on their operating width, the source and destination locations (immediate, register, and memory) and group them together. 

For example in x86, there are about $200$ different \texttt{MOV} instructions operating on different bit width, sources, and destinations, like \texttt{MOV32r0}, \texttt{MOVZX64rr16}, \texttt{MOVAPDrr}, etc. All such opcodes are grouped together as a generic \texttt{MOV} token while forming the triplets. 
The obtained triplets are fed to the TransE model to generate the embeddings for each entity-resulting in seed embedding vocabulary.

\paragraph{Representing instructions}
For a given MIR instruction with opcode $O$ and $n$ arguments $A_1, A_2, \dots , A_n$, its representation is computed as 
\[
    W_o\cdot\llbracket \mathbf{O} \rrbracket + W_a\cdot\left(\llbracket \mathbf{A_1} \rrbracket + \llbracket \mathbf{A_2} \rrbracket + \cdots + 
    \llbracket \mathbf{A_n} \rrbracket \right), W_o > W_a 
\] 
where $W_o, W_a \in (0,1]$ correspond to the weights of opcodes and arguments, $\llbracket \cdot\rrbracket $ denotes the embedding of the entity from seed-embedding vocabulary, and operators $\cdot$ and $+$ denote multiplication and vector addition respectively.

\paragraph{Interference Graphs}
As mentioned earlier, the MIR at the stage of register allocation contains partial physical register assignments and virtual registers. The physical registers are assigned for the instruction operands that have restrictions on the particular register to be used. Virtual registers are used in all other places.
Consequently, we need to take into account the edges corresponding to both virtual and physical registers. Virtual registers are marked with the register class so that assignments can only be one among the physical registers in that class.

For computing $G$, considering the interferences between the (physical $\longleftrightarrow$ virtual) and (virtual $\longleftrightarrow$ virtual) registers is sufficient as $G$ is bidirectional and we do not need to worry about the physical registers that are already assigned. 

We use a collection of instructions in the live-range of a variable to represent a vertex of the interference graph. 
Each instruction is \textit{represented} in $\mathbb{R}^n$ using \mirtovec{} embeddings. 
Consequently, a vertex $v$ is represented as a matrix of embeddings $\llbracket v \rrbracket$ in $\mathbb{R}^{m \times n}$, where $m$ denotes the number of instructions in its live range.

Gated Graph Neural Networks (GGNNs) are widely used in programming language modeling~\citep{cummins2021PrograMl,mendis2019ImitationLearning} especially when the inputs are modelled as graphs.
GGNNs involve message passing between the nodes of the graph. Information propagates across multiple nodes to arrive at the representation for a given node.
Also, they allow annotating the nodes and edges based on their types and properties, and consider these while learning a representation.
We use GGNNs to process the embedded interference graph to get the final representation. This network transforms $\mathbb{R}^{m \times n} \rightarrow \mathbb{R}^{d}$.
We set $d = n$ in our experiments and annotate the graph with the following node types, along with spill weights.
\begin{itemize}
    \item \textbf{Not visited} --- nodes that are not visited yet.
    \item \textbf{Spill} --- nodes that are marked as spill.
    \item \textbf{Colored} --- nodes that are assigned a register.
\end{itemize}

Such node representations are propagated through a GGNN by means of message passing. 
Messages received from adjacent nodes are aggregated and passed through a Gated Recurrent Unit~\citep{cho2014GRU} to yield a final representation.

\section{Compiler Integration}

The integration of DL/RL models into optimizing compilers can be a hurdle, for both research and practical deployment.
Problems like vectorization~\cite{hajali2019neurovectorizer} or phase ordering~\cite{Jain-ISPASS-22-DBLP:conf/ispass/JainAVU22} are adeptly controlled through optimization flags, carrying information from the predictions of the model to the compiler.
Register allocation however poses an inherent difficulty as the ML-decisions and the optimization algorithm are deeply intertwined; the model predicts the final allocation based on the compiler generated code from its splitting decisions. One na{\"i}ve approach is to rely on Python bindings for integration; however, this involves large overhead.

We propose \llvmgrpc{}, a novel infrastructure for efficient communication between the Python model and C++ compiler to support \emph{both training and inference}. 
\llvmgrpc{} involves gRPC calls~\cite{grpc} with the LLVM toolchain, leveraging its modular structure as an LLVM library. This gives the end-user the option of designing custom RPC calls that can operate on any of the module, basic-block, loop or function of the input program.  \llvmgrpc{} allows bi-directional communication between ML models and the compiler, during both training and inference. To our knowledge, this facility is not available in other frameworks. 

The splitting decision by the model is communicated to the compiler via \llvmgrpc{}, which then applies it and responds back with the update containing new interferences and live ranges.
The model then updates the interference graph using the received information and continues the traversal. 
After processing all vertices of $G$, all the coloring decisions are communicated to the compiler as a color map. 

\begin{table}[h!tb]
        \small
        \caption{Allocatable Registers in x86 and AArch64}
        \begin{tabular}{l@{\;}l}
        \toprule
        \textbf{Arch.} & \textbf{Registers} \\
        \midrule
        x86 & [A-D]L, [A-D]X, [E,R][A-D]X, [SI,DI]L, [E,R][SI,DI], \\
            &  SI, DI, R[8-15][B,W,D], FP[0-7], [X,Y,Z]MM[0-15]\\
        AArch64 &  [X,W][0-30], [B,H,S,D,Q][0-31]\\
        \bottomrule
    \end{tabular}
    \label{tab:registers}
\end{table}

\begin{figure*}[h!tb]
  \resizebox{1\textwidth}{!}{%
  \begin{minipage}[b]{\textwidth}
  \begin{minipage}[t]{.625\textwidth}
    \centering
    \includegraphics[scale=0.47]{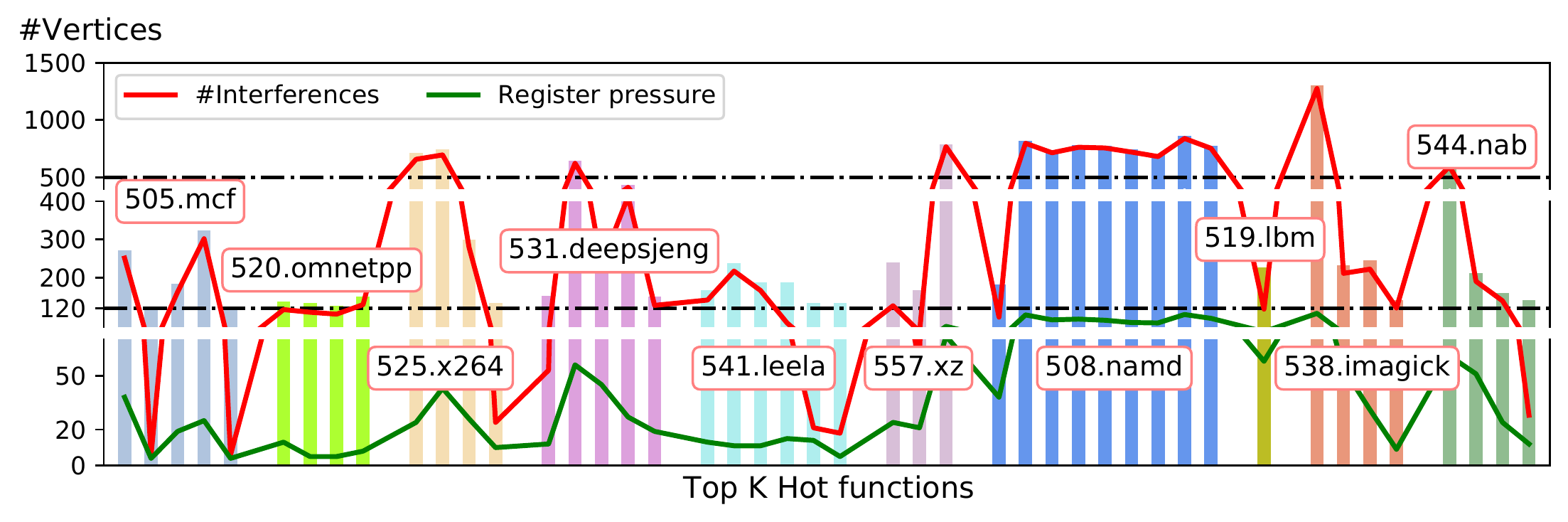}
    \caption{Correlation between \#Vertices, \#Interferences and Register Pressure}
    \label{fig:spec17-characterization}
  \end{minipage}
  \begin{minipage}{.37\textwidth}
        \small
        \vspace{-3.3cm}
        \captionof{table}{\% runtime diff. over \rabasic on hot functions}
        \label{tab:x86-hotfns}
        \resizebox{1\textwidth}{!}{%
            \small
            \addtolength{\tabcolsep}{-2pt}  
            \begin{tabular}{lrrr|rrr}
            \toprule
              \multirow{3}{*}{} & \multicolumn{3}{c}{\textbf{SPEC CPU 2017}} & \multicolumn{3}{c}{\textbf{SPEC CPU 2006}} \\
             & \multirow{2}{*}{\hskip-2em\ragreedy\!} & \multicolumn{2}{c}{\textbf{\toolname{}}} & \multirow{2}{*}{\hskip-1ex\ragreedy\!} & \multicolumn{2}{c}{\textbf{\toolname{}}} \\
             & \textbf{} &\multicolumn{1}{c}{\textbf{L}} & \multicolumn{1}{c}{\textbf{G}} & \textbf{} & \multicolumn{1}{c}{\textbf{L}} & \multicolumn{1}{c}{\textbf{G}} \\
             \midrule
            Average                   & 6.2  & \textbf{7.3}  & 4.8 & \textbf{-1.5} & -2.1 & -1.6 \\
            \rlap{\# (val\textgreater 0)} & \textbf{23} & \textbf{23} & 17 & 16 &	\textbf{17} &	13  \\
            \rlap{\# (val\textless 0)}    & \textbf{8}  & \textbf{8}  & 14 & 19 & \textbf{18} & 22 \\
            Max                       & 44.0 & \textbf{44.4} & 41.3  & \textbf{12.7} & 10.4 & 6.2 \\
            Min                       & -7.7 & \textbf{-4.4} & -10.8 & -51.4 & -52.5 & \textbf{-13.1} \\
            \bottomrule
            \end{tabular}
            \addtolength{\tabcolsep}{1pt}  
        }
  \end{minipage}
  \end{minipage}
  }
\end{figure*}

\section{Experimental Evaluation}
\label{sec:experiments}

We first discuss the experimental setup, followed by a characterization of the benchmarks. Then, we report the results on x86, followed by the results on AArch64 and an explanation of the results.
Finally, we report improvements on the regression cases using policy improvements.

\subsection{Setup}
\label{sec:experimental-setup}
For training \mirtovec{} representations, we randomly select 2K source files from SPEC CPU 2017 benchmarks and C++ Boost library.
MIR triplets are generated by applying \texttt{-O3} optimization flag.
The seed embedding vocabulary is obtained by training a TransE model~\cite{transe-Bordes:2013:TEM:2999792.2999923}
on generated triplets, by running an SGD optimizer over 1000 epochs to obtain an embedding vector of 100 dimensions. 
We obtain 1 Billion MIR triplets from which $\{675, 315\}$  entities and $\{25, 17\}$ relations are generated for $\{\textrm{x86}, \textrm{AArch64}\}$ respectively. 

We target a complex x86 (Intel Xeon SkyLake W2133, 6 cores, 32GB RAM), and a simpler mobile AArch64 (ARM Cortex A72, 2 cores, 8GB RAM) processors. 
We consider allocations of general purpose, vector, floating point registers for both x86 and AArch64 (listed in Tab.~\ref{tab:registers}); other registers like \texttt{eflags} are pre-assigned before any \regalloc{}. 

Our framework is implemented as a pass \mlregalloc{} in LLVM~10.0.1,
using gRPC~v1.34.
We train the RL models using the PPO policy with the standard set of hyperparameters on the training set of functions selected from SPEC CPU 2017, until convergence of reward graph. 
There were about 11K and 30K functions (with 120--500 vertices) in SPEC CPU 2017 benchmark suite in x86 and AArch64 respectively, out of which we choose 5K functions at random (from SPEC 2017) for training.
Training was done on a 32GB Tesla V100 GPU and sampling was done using 12 threads of a server with Intel Xeon Platinum 8168 processors.

\paragraph{Model Architecture}
Each agent learns a policy depending on its model architecture.
For GGNN, we use two fully connected (FC) layers to normalize the input, followed by a RNN layer for message passing. For the agents, we use simple neural networks with FC layers: node selector and splitter use four FC layers each, while task selector and coloring agents use three FC layers each with batch normalization. 
Everywhere, ReLU is used as the activation function.

\paragraph{Benchmarks}
In our experiments on x86, we consider C/C++ benchmarks with less than 1 MLOC (1 Million Lines of Code) from SPEC CPU 2006 and 2017. This constitutes (8 Int + 5 FP) 13 benchmarks in SPEC CPU 2006, and  (8 Int + 5 FP) 13 benchmarks in SPEC CPU 2017 benchmark suites. We were able to successfully compile the benchmarks listed in Tab.~\ref{tab:x86-runtime}. 
We observed $4$ compilation errors, and $4$ runtime errors due to the engineering and integration issues.

\subsection{Characterization of Benchmarks}

Let us now characterize the SPEC CPU 2017 benchmarks for choosing the graphs for experimentation. We study the $45$ \textit{hot} functions (profiled with the \texttt{perf} tool~\cite{perf}) that take at least 5\% of the total execution time of a benchmark.
The number of vertices in the interference graphs of these hot functions along with the number of interferences and register pressure is shown in Fig.~\ref{fig:spec17-characterization}. 
We compute register pressure as the maximum number of overlapping live ranges across all program points of a function.

It can be observed that out of $45$ hot functions, $44$ have more than 120 vertices in their interference graphs, while $31$ (the majority) of them have between $120$ and $500$ vertices. 
It can be also be seen that the graph size has a strong correlation with the number of interferences and the register pressure. Meaning, the vertices and edges show linear correlation (not quadratic), and the vertices and register-pressure also show near linear correlation (not quadratic)~\cite{bouchez2006register}.

As the number of vertices and the register-pressure are positively correlated with each other, in general the graphs with higher number of vertices are harder to allocate. 
Hence, we consider the functions that have at least $120$ vertices for allocation through \toolname{}.  
We limit the maximum number of vertices to be $500$, as most of the hot functions are in this range ($120$--$500$), and also for ease of training.
Functions that are not in this range are processed using \ragreedy.

\newcommand{\first}[1][]{\cellcolor{sgreen}}
\newcommand{\second}[1][]{\cellcolor{yellow}}

\begin{table}[h!tb]
\small
\centering
\caption{Runtime improvement (s) on x86 over \rabasic{}, highlighting the
\colorbox{sgreen}{Highest} and \colorbox{yellow}{Second highest} improvements.}
\label{tab:x86-runtime}
\begin{tabular}{lrrrrr}
\toprule
\multicolumn{1}{c}{\multirow{3}{*}{\textbf{Benchmarks}}} &
  \multicolumn{1}{c}{\multirow{3}{*}{\textbf{\begin{tabular}[c]{@{}c@{}}\!\!Runtime\!\!\\ \rabasic\end{tabular}}}} &
  \multicolumn{4}{c}{\textbf{Diff.\ from \rabasic (\rabasic - x)}} \\
\cmidrule(lr){3-6}
\multicolumn{1}{c}{} &
  \multicolumn{1}{c}{} &
  \multicolumn{1}{c}{\multirow{2}{*}{\rapbqp}} &
  \multicolumn{1}{c}{\multirow{2}{*}{\ragreedy}} &
  \multicolumn{2}{c}{\textbf{\toolname{}}} \\
\cmidrule(lr){5-6}
\multicolumn{1}{c}{} &
  \multicolumn{1}{c}{} &
  \multicolumn{1}{c}{} &
  \multicolumn{1}{c}{} &
  \multicolumn{1}{c}{\textbf{L}} &
  \multicolumn{1}{c}{\textbf{G}} \\
\midrule
401.bzip2              & 360.6       & -7.3  & \second 7.5   & -1.1  & \first 10.8 \\
429.mcf                & 233.8       & \second 1.4   & -2.9  & \first 2.7   & -3.6  \\
445.gobmk              & 322.3       & -3.3  & \first 6.4   & \second 2.4   & 1.7   \\
456.hmmer              & 284.3       & 1.8   & \first 6.1   & \second 5.0   & -37.6 \\
462.libquantum         & \first 256.4       & -10.1 & \second -1.1  & -2.2  & -6.7  \\
471.omnetpp            & 305.7       & \second 0.7   & \second 0.4   & \first 1.2   & \first 1.2   \\
433.milc               &\first 349.1       & -16.6 & \first 0.1   & -13.8 & \second -7.0  \\
470.lbm                & 184.0       & -7.9  & \first 3.0   & \second 2.3   & 1.4   \\
482.sphinx3            & 366.0       & -37.5 & \first 1.6   & -3.1  & \second -2.7  \\
\midrule
505.mcf\_r             & 344.9       & \second 4.5   & -1.6  & \first 8.6   & -4.7  \\
520.omnetpp\_r         & 475.7       & \first 6.4   & \first 6.4   & \second 2.4   & \second 2.8   \\
531.deepsjeng\_r       & 299.9       & 4.6   & \first 16.0  & 9.9   & \second 12.8  \\
541.leela\_r           & 439.5       & 1.6   & \first 7.1   & 0.4   & \second 1.9   \\
557.xz\_r              & 371.5       & -0.6  & \second 11.9  & \first 12.1  & -8.5  \\
508.namd\_r            & 236.5       & 2.5   & \first 23.5  & 9.1   & \first 23.8  \\
519.lbm\_r             & 261.8       & 1.4   & \first 57.7  & \second 50.9  & \first 58.1  \\
538.imagick\_r         & 479.3       & -16.9 & \second 115.5 & \first 118.8 & \first 118.4 \\
544.nab\_r             & 417.5       & 5.8   & \second 132.1 & \second 131.3 & \first 134.4 \\
\midrule
\multicolumn{2}{r}{\textbf{Average}} & -3.9  & 21.7  & 18.7  & 16.5 \\
\bottomrule
\end{tabular}
\end{table}
\quad
\begin{table}[h!tb]
\small
\caption{\% speedup of \ragreedy and \toolname{} over \rabasic on SPEC 2006 and 2017 (x86)}
\centering
\label{tab:top5-best-worst-fns}
\addtolength{\tabcolsep}{-2pt}    
\begin{tabular}{llrrr}
\toprule
  \multicolumn{1}{c}{\textbf{B/M}} & \multicolumn{1}{c}{\textbf{Functions}} & \ragreedy & \toolname{} & \textbf{Diff.} \\
  \midrule
 \multicolumn{5}{c}{\underline{Top 5 functions with highest \% speedup (over \ragreedy)}}\\
  401  & BZ2\_compressBlock	& -51.3	& -5.2	& 46.1\\
  445 & do\_get\_read\_result	& -12.0	& -0.5	& 11.5\\
  482 & mgau\_eval	& -6.0	& 0.3	& 6.3 \\
  429 & price\_out\_impl	& -0.8	& 2.3 & 3.2 \\
  445 & subvq\_mgau\_shortlist	& -9.8	& -6.9	& 2.9 \\
  \midrule
    538 & GetVirtualPixelsFromNexus            & 8.3       & 28.8  & 20.4\\
    538 & SetPixelCacheNexusPixels            & 4.7       & 21.9  & 17.2\\
    505 & cost\_compare	                    & -7.7	     & 8.1	& 15.8 \\
    557 & lzma\_mf\_bt4\_skip	                & -1.8	     & 3.63	& 5.5 \\
    525 & biari\_decode\_symbol	                & -2.7	     & 2.7	& 5.4 \\
\midrule[1pt]
\multicolumn{5}{c}{\underline{Top 5 functions with highest \% slow-down (over \ragreedy)}}\\
  456 & P7Viterbi	& 2.2	& -13.1	& -15.3\\
  482 & vector\_gautbl\_eval\_logs3	& 11.9	& -2.5	& -14.4\\
  401 & mainGtU	& 0.3	& -9.6	& -10.0\\
  401 & fallbackSort	& 12.6	& 6.2	& -6.4\\
  445 & fastlib	        & 4.8	& -1.1	& -5.9 \\
\midrule
  557 & lzma\_mf\_bt4\_find	                & 1.5	      & -10.7	& -12.3\\
  531 & feval                            	& 26.4	& 17.7	    & -8.6\\
  505 & primal\_bea\_mpp	                & 0.9	& -7.6      & -8.5\\
  541 & FastBoard::self\_atari	            & 3.7	&  -0.1 & -5.8\\
  541 & qsearch                             & 6.6	& 1.5 & -5.0 \\
\bottomrule
\end{tabular}
\addtolength{\tabcolsep}{1pt}
\end{table}

\subsection{Runtimes on x86}

We study the SPEC CPU 2017\&2006 benchnmarks and compare the results with LLVM's allocators. \rabasic, \ragreedy and \rapbqp generally outperform the \rafast allocator. However, there is no single allocator among these three that perform the best for \textit{all} programs.
Hence, we compare our results with these three allocators.

In Tab.~\ref{tab:x86-runtime}, we show the runtimes obtained by \rabasic and the improvements obtained over it by other allocators for each benchmark.  These are obtained by taking the median of three executions. Positive (negative) numbers indicate speedups (slow-downs) over \rabasic. 
For \toolname{}, we train two models: 
one trained only using local rewards (L) and another with local along with the global reward (G).

On average, \rlforreall (\rlforreallg) yields about $19s$ ($17s$) improvement over \rabasic{}; \ragreedy results in an improvement of about $22s$.
\rlforreall (\rlforreallg) shows speedup over \rabasic{} in $14$ ($11$) out of $18$ benchmarks.
The highest and second highest improvements in runtimes over \rabasic are highlighted in Tab.~\ref{tab:x86-runtime}. In particular, \toolname{} (L or G) results in highest or second highest improvements over \rabasic in $17$ out of $18$ benchmarks. 
As it can be seen, the runtimes obtained by our framework are very close to \ragreedy. 
In comparison, \rlforreall{} (\rlforreallg{}) results in an improvement on $5$ ($6$) benchmarks over \ragreedy{}.
And those with slow-downs, runtimes of $12$ ($11$) benchmarks are within $1\%$ of \ragreedy{}, and only $1$ show more than $4\%$ slow-down. 

To obtain confidence intervals, we ran benchmarks $8$ times and observed that the noise was under 1\% consistently across all benchmarks for a 95\% confidence interval, except for \texttt{libquantum}, where the noise was  1.8\% (1.2\%) in \rlforreall{} (\rlforreallg{}).
We also have empirical results on numerical kernels. On PolyBench~\cite{pouchet-polybench} benchmark, our results show similar performance: \toolname{} obtains an average runtime of $43.5s$ ($3.6s$) in comparison $43.6s$ ($3.6s$) obtained by \ragreedy{} on Extra-Large (Large) input size.

\paragraph{Analysis of Hot functions}
We did a study at function level, focusing on the hot functions. There were 35 and 31 allocated hot functions in SPEC 2006 and 2017. In Tab.~\ref{tab:x86-hotfns}, we show the percentage difference in runtime improvements obtained by \ragreedy{} and \toolname{} allocators in comparison to \rabasic{}. It can be seen that \toolname{} results in improvements on largest number of functions, and minimum number of slow-downs over \rabasic.

On average, in SPEC 2017 \rlforreall{} improves over \rabasic{} by $7\%$, while \ragreedy{} results in a similar improvement of about $6\%$.
When it comes to SPEC 2006, to our surprise, \ragreedy did not show improvement on average runtime  among the hot functions.
It is mainly due to a $51.3\%$ slow-down observed on \texttt{BZ2\_compressBlock} from \texttt{Bzip2} benchmark. 
In comparison, \rlforreallg{} results in a lesser slow-down of about $5\%$ on this function.

In \texttt{imagick} benchmark, \toolname{} (both L and G) obtains improvements over \ragreedy{}, on all the three allocated hot functions:
On \texttt{GetVirtualPixelsFromNexus} and \texttt{SetPixel- CacheNexusPixels} functions, \rlforreal{} improves by over $29\%$ and $22\%$, where \ragreedy{} results in an improvement of about $8\%$ and $5\%$; on \texttt{MeanShiftImage} function, \ragreedy and \toolname{} show a similar improvement of about $44\%$. 
We list the top $5$ hot functions that show highest \%~speedup and \%~slow-down in comparison to \ragreedy from both SPEC 2006 and 2017 in Tab.~\ref{tab:top5-best-worst-fns}.

\subsection{Runtimes on AArch64}
In this section, we study the performance of SPEC CPU 2017\&2006 benchmarks on AArch64. 
As mentioned earlier, the runtimes shown correspond to the median of three runs. For the cases where there were significant differences in runtimes (about $20s$) across different runs, we take a median of five runs. 
We cross-compile the binaries from x86 targetting the AArch64 board by running the inference. 
We skipped the benchmarks like \texttt{perlbench}, \texttt{h264ref}, \texttt{xalancbmk} and \texttt{sphinx3}, as they are known to have cross-compilation issues, or they fail on compilation/execution with the standard register allocators of LLVM. 
The runtimes obtained by \rabasic{}, the improvements obtained over it by other allocators and \rlforreallg{} are listed in Tab.~\ref{tab:aarch64-runtime}. 
On average, \rlforreal{} achieves an improvement of $18s$, whereas \ragreedy{} achieves an improvement of about $19s$. Also, \rlforreal{} achieves highest or second highest improvements in all the benchmarks except four.
These results demonstrate that the learned heuristics from our model work well on different architectures.

\begin{table}[tb]
\caption{Improvement in runtimes (s) on AArch64 over \rabasic allocator. 
\colorbox{sgreen}{Highest} and \colorbox{yellow}{Second highest} improvements are highlighted.}
\label{tab:aarch64-runtime}
\small
\begin{tabular}{lrrrr}
\toprule
\multicolumn{1}{c}{\multirow{2}{*}{\textbf{Benchmarks}}} &
  \multicolumn{1}{c}{\multirow{2}{*}{\textbf{\begin{tabular}[c]{@{}c@{}}Runtime\\ \rabasic\end{tabular}}}} &
  \multicolumn{3}{c}{\textbf{Diff.\ from \rabasic (\rabasic - x)}} \\
  \cmidrule(lr){3-5} 
\multicolumn{1}{c}{} &
  \multicolumn{1}{c}{} &
  \multicolumn{1}{c}{\multirow{1}{*}{\textbf{\rapbqp}}} &
  \multicolumn{1}{c}{\multirow{1}{*}{\textbf{\ragreedy}}} &
  \multicolumn{1}{c}{\multirow{1}{*}{\textbf{\toolname{}}}} \\
\midrule
401.bzip2 &  1366.9 &   -41.1 &   \first 15.6  & \second 12.8 \\
429.mcf &   \second 1320.5 &   -12.7 &   -7.5 & \first 1.6\\
445.gobmk &   992.8 &    \second 15.6 &  \first 26.1  &  \second 14.5 \\
462.libquantum &   1627.6 &   -8.7 &   \second 4.5  & \first 9.6\\

433.milc &   1251.1 &  \second 59.2 &   \first 70.9  &  45.4\\
444.namd &   855.3 &   2.7 &   \first 21.8  &  \second 18.8\\
470.lbm &   1604.3 &   \second -6.4 &   -16.6  & \first 16\\
\midrule
505.mcf\_r & 1535.1 & \first 25.9 & \second 1.9 & -12.8\\
508.namd\_r & 845 & 0.4 & \second 34.5 & \first 40.1 \\
523.xalancbmk\_r &   979.1 &   \first 8.1 &   -3.4  & \second 4.4 \\
531.deepsjeng\_r &   777.2 &   \second 10.0 &   \first 30.5 & 4.5\\
541.leela\_r &   \first 1067.9 & \second -11.3 &   \first -0.1 & -19.5 \\
557.xz\_r &   1163.2 &   \second 3.7 &   \first 22.2 & \first 21.3\\
519.lbm\_r &   1657 &   \first 50.9 &   -1.6  & \second 39.8\\
538.imagick\_r &   1244.5 &   -3.9 &   \first 75.8 & \second 65.6\\
544.nab\_r &   \second 1170.7 &   -7.7 &   \first 31.5  &  \first 32.4\\
  \midrule
\multicolumn{2}{r}{\textbf{Average}} &   5.3 & 19.1 & 18.4\\
 \bottomrule

\end{tabular}
\end{table}

\subsection{Policy Improvement on Regression Cases}
In traditional compilers, heuristic tuning for optimization is an iterative process: human experts identify regression cases, and heuristics are tuned to identify 
 cases of regression.
Similar to this spirit, Trofin et al.~\cite{trofin20MLGO} (MLGO), propose a \textit{policy improvement cycle}, where the learned RL policy is fine-tuned to check if it fares well on the regression cases. 

In this section, we attempt to tune the learned policy to evaluate if the regression cases can be improved.
For this purpose, we identified poorly performing benchmarks from each configuration: \texttt{milc} on \rlforreall{}, \texttt{hmmer} and \texttt{xz} on \rlforreallg{}. 
The learned model is then retrained (fine-tuned) on the hot functions of these benchmarks.
Upon training, we observe a positive improvement on all three regression cases: $12s$ on \texttt{milc}, $10s$ and $6s$ on \texttt{hmmer} and \texttt{xz} benchmarks. 
This experiment makes a strong case for online or continuous learning~\cite{alpaydin2020introduction}, where the learning continues during deployment for betterment of policy.

\subsection{Discussion}
We demonstrate performance results \textit{on par with the best allocators} currently available in LLVM: RL4ReAl is \textit{most frequently the best or second best allocator}, and there is \textit{no single allocator} that performs best across all benchmarks (see Tables~\ref{tab:x86-runtime} and ~\ref{tab:aarch64-runtime}). It is well known that register allocation is one of the hard compiler optimization problems, and the baseline heuristics achieve excellent results that cannot be easily improved upon in terms of wall-clock time. For example, the studies of Pereira et al.~\cite{Pereira-PLDI-08-10.1145/1379022.1375609}, Shin et al.~\cite{Shin-LCPC21-10.1007/978-3-030-99372-6_3}, Kim et al.~\cite{kim2022pbqp}, and the report on the PBQP solver~\cite{pbqp-llvm} were \textit{not able to} significantly outperform baseline heuristics across benchmarks, and they report performance numbers in the \textit{same ballpark} as the ones that we obtain. 

In this work, we consider the major sub-tasks/strategies of register allocation: coloring, splitting and spilling, with a focus on building the first end-to-end RL model integrated with LLVM. As mentioned earlier, \ragreedy also admits register coalescing as one of its strategies. 
We consider coalescing along with other possible strategies like multi-allocation, register packing, spilling to vector registers, as possible incremental extensions for a future work. 
These additional strategies could result in further runtime improvements.
It could however be noted that even without admitting these additional strategies---and just relying on the ones that are available in LLVM---\toolname{} gives competitive numbers vs. the state-of-the-art regallocs in LLVM.

It can be noted that these results have been obtained fully automatically, against production-grade allocators tuned over many man-decades of experience and effort.
\section{Related Work}

Recently, several ML-supported compilers were proposed, leveraging representation learning techniques for compiler optimizations \cite{hajali2019neurovectorizer, IthemalMendis19a, Mammadli2020, Jain-ISPASS-22-DBLP:conf/ispass/JainAVU22}.
These works use learned embeddings like inst2vec \cite{ncc}, IR2Vec \cite{VenkataKeerthy-2020-IR2Vec}, Flow2Vec \cite{flow2vec2020} for representing the input programs to the ML model.
We model a complex register allocation problem using RL and propose MIR2Vec to represent programs in MIR form.

An initial attempt to solve \regalloc{} using ML models by Das et al.~\cite{das2020} uses an LSTM to come up with an initial coloring scheme; it undergoes a \textit{correction phase} to rectify the inconsistency in coloring interferences.
Their work focuses on the graph coloring problem, and to our understanding, the solution was not integrated to obtain the final register assignments.  Another recent work by Kim et al.~\cite{kim2022pbqp} proposes an RL-based solution inspired from AlphaZero~\cite{HuangAlphaGoZero2019} for solving PBQP constraints by reducing the search space; they use Monte Carlo Decision Trees to simulate solving the PBQP graphs. 
Their model focuses on an irregular and custom architecture for Automated Test Equipments.  \rlforreal{} is the first end-to-end application of RL for solving the generic \regalloc{} problem; does not need a separate correction phase, and is integrated as a \mlregalloc{} pass in LLVM, and reports results that are comparable to the regallocs in LLVM.

Compiler-Gym~\cite{cummins2021compilergym} is a recent approach designed to leverage Python libraries for solving compiler optimization problems; it exposes RL environments and datasets for training. However it currently does not support integrating these trained models in the compiler pipeline for both training or deployment.
Another framework, MLGO~\cite{trofin20MLGO} integrates trained ML/RL models within the LLVM compiler. For this purpose, the compiler loads a trained model and accesses it via C++ APIs of Tensorflow or ahead-of-time generated code (release mode).
The framework is used in production, with improved decisions for inlining for size, and live-range eviction (in \regalloc{}) when compared to the compiler's default heuristics.
MLGO addresses a narrower space of the register allocation problem, but its deep integration of a precompiled ML model into LLVM hints at a path for further integration of deployment of our approach in a production compiler.

\section{Conclusion}

We propose a target-independent Reinforcement Learning approach to the Register Allocation problem.
We use a multi-agent hierarchical algorithm to learn a policy for three of the main sub-tasks of register allocation, including coloring, live range splitting, and spilling. Semantic correctness is ensured by the constraints encoded as the action masks for the agents.
Our method often exhibits better allocations and generally perform on-par with the standard register allocators of LLVM.
\rlforreal{} opens up new opportunities for research on \regalloc{} and on other backend compilation problems.
Source code and the related artifacts are available in \url{https://compilers.cse.iith.ac.in/research/rl4real}.

\begin{acks}
We are grateful to Govindarajan Ramaswamy, Dibyendu Das, Yundi Qian, Mircea Troffin, Yanqi Zhou and Nilesh Shah for valuable discussions and feedback. 
We would like to thank Utpal Bora, Sayan Dey, Shikhar Jain, Soumya Banerjee and Raj Ambekar for their fruitful comments and help in validating and improving the artifacts of this work.
We acknowledge National Supercomputing Mission (NSM) for providing \textit{PARAM Seva} computing resources at IIT Hyderabad and HPCE, IIT Madras.
This work is partially funded by a Google PhD fellowship, an NSM research grant (MeitY/R\&D/HPC/2(1)/2014), and a faculty research grant from AMD.
\end{acks}

\bibliographystyle{ACM-Reference-Format}

\balance

\bibliography{references}

\end{document}